\documentclass[fleqn,10pt]{SelfArx} 

\pdfoutput=1
\usepackage{graphicx}
\usepackage[english]{babel} 
\usepackage{nicefrac}
\usepackage{subfigure}
\usepackage{enumitem} 
\usepackage{eucal}
\usepackage{pbsi}
\usepackage[linesnumbered,ruled,vlined]{algorithm2e}

\newcommand{\params}[1]{\left( #1\right) }
\usepackage{pict2e}

\newcommand{\Mod}[1]{\ \mathrm{mod}\ #1}
\newcommand{\set}[1]{\{ #1 \}}
\newcommand{\partialfx}[2]{\nicefrac{\partial #1}{\partial #2}}
\newcommand{\partialFx}[2]{\frac{\partial #1}{\partial #2}}
\newcommand{\etal}{et\,al. }
\newcommand{\G}[1]{\enspace--\,#1\,--\space}
\newcommand{\g}[2]{\enspace--\,#1#2}
\newcommand{\parr}[1]{\vspace{2mm}\noindent \textit{#1.} \\}
\newcommand{\pp}[2]{\textit{\small{$P(U=$ `#1' $)=#2 $}}}

\definecolor{color1}{RGB}{0,0,90} 
\definecolor{color2}{RGB}{0,20,20} 

\usepackage{hyperref} 
\hypersetup{hidelinks,colorlinks,breaklinks=true,urlcolor=color2,citecolor=color1,linkcolor=color1,bookmarksopen=false,pdftitle={Title},pdfauthor={Author}}

\JournalInfo{Journal, Vol. XXI, No. 1, 1-5, 2013} 
\Archive{Additional note} 

\PaperTitle{A Geometric Approach to Harmonic Color Palette Design} 

\Authors{Carlos Lara-Alvarez\textsuperscript{1}*, Tania Reyes\textsuperscript{2}} 
\affiliation{\textsuperscript{1}\textit{CONACYT Research Fellow -- Center for Mathematical Research CIMAT, Zacatecas, Mexico}} 
\affiliation{\textsuperscript{2}\textit{Software Engineering Group, Center for Mathematical Research CIMAT, Zacatecas, Mexico}} 
\affiliation{*\textbf{Corresponding author}:}

\newcommand{\keywordname}{Keywords} 

\Abstract{  
	We address the problem of finding harmonic colors, %
	this problem has many applications, from fashion to industrial design. %
	The proposed approach enables to evaluate and generate color combinations incrementally. %
	In order to solve this problem we consider that colors follow normal distributions in tone (chroma and lightness) and hue. The proposed approach relies in the CIE standard for representing colors and evaluate proximity. %
	Other approaches to this problem use a set of rules. Experimental results show that lines with specific parameters \G{angles of inclination, and distance from the reference point}  are preferred over others, and that uncertain line patterns outperform non-linear patterns.
}

\begin{document}

\flushbottom 

\maketitle 


\thispagestyle{empty} 


\section{Introduction} 

\addcontentsline{toc}{section}{Introduction} 

At first the clothing was created to protect people form weather an other uncomfortable conditions, but it has evolved to become a form of identity, expression, and creativity. Clothing is an essential part in the social perception; i.e., how people form impressions of and make inferences about other people. For instance, several clothing categories have been created for different ocassions \g{e.g., for formal events, work, and sport}{.} In this sense, color of clothing is a key factor that mainly drives these people impressions. Colors seen together to produce a pleasing affective response are said to be in harmony \cite{burchett2002color}. Color harmony represents a satisfying balance or unity of colors. The study of harmonious colors has a long tradition in many areas, and it is attractive for artists, industrial applications, and scientists in different fields.

We envisage a color harmony model that integrates several divergent theories described in the literature. For this purpose, colors are represented in the CIE L*C*h \G{a.k.a. CIELCh} color scale. This scale is approximately uniform in a polar space with Ligthness (L), chroma (C) and hue (h), being the height radial and azimuthal coordinates respectively.

\begin{figure}[b!]
	\begin{center}
		\def\svgwidth{0.9\columnwidth}\input{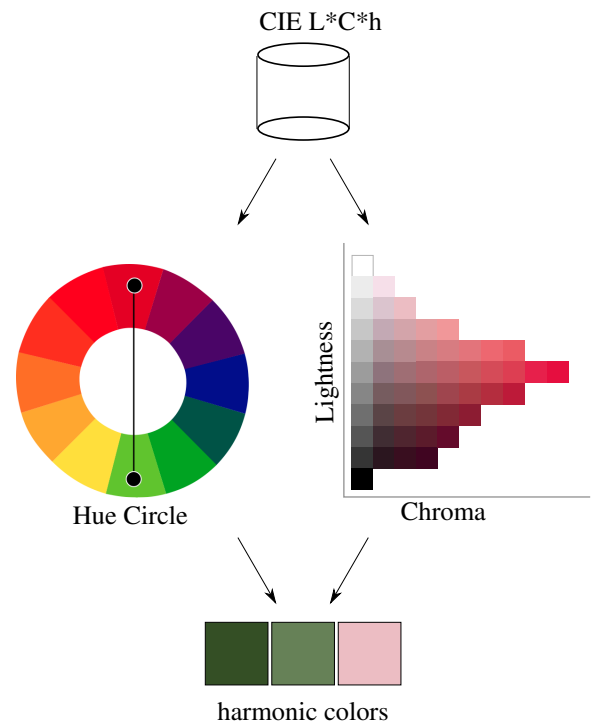}
	\end{center}
	\caption{Color harmony models should follow patterns both in the color wheel and in the tone plane. This paper studies harmonic colors in the tone \mbox{\G{chroma-lightness}} plane.} 
	\label{fig:thispaper}
\end{figure}

Like other approaches, the proposed one considers that harmonious colors can be identified from the analysis of two components (Fig. \ref{fig:thispaper}):
\begin{enumerate}[label=(\roman*)]
	\item relationship of  colors in the color wheel, and
	\item  relationship of colors in the plane of tones (chroma--lightness). 
\end{enumerate} 

This paper proposes a generalization of geometric approaches to color harmony using a normal distribution to describe color variability. Regarding the color relation in the hue circle, the proposed approach uses conventional techniques to find harmonious hue patterns. These patterns\G{adjacency, opposite, triad} are very common in the literature. On the other hand, the proposed approach considers that harmonious colors follow specific paths in the plane of tones, this paper explores the hypothesis that harmonic colors follow straight lines in the tone plane. We state that many theories can be explained by describing color variability in a proper manner; for instance, a simple straight line with uncertainty describes many patterns proposed by Matsuda \cite{matsuda1995color}.

\begin{figure}[b!]
	\begin{center}
		\def\svgwidth{0.5\columnwidth}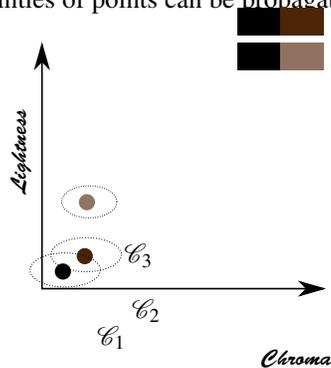
	\end{center}
	 	\caption{Color harmony in the tone plane:  colors $\{\CMcal{C}_1, \CMcal{C}_2\}$ are not harmonic because their distance is close to zero; on the other hand,  $\{\CMcal{C}_1, \CMcal{C}_3\}$ y $\{\CMcal{C}_2, \CMcal{C}_3\}$ are harmonious.}
	\label{fig:twocolors}
\end{figure}
\vspace{0.3cm}

\begin{figure}[t!]
	\centering
\subfigure[]{
		\centering
	\def\svgwidth{0.45\columnwidth}		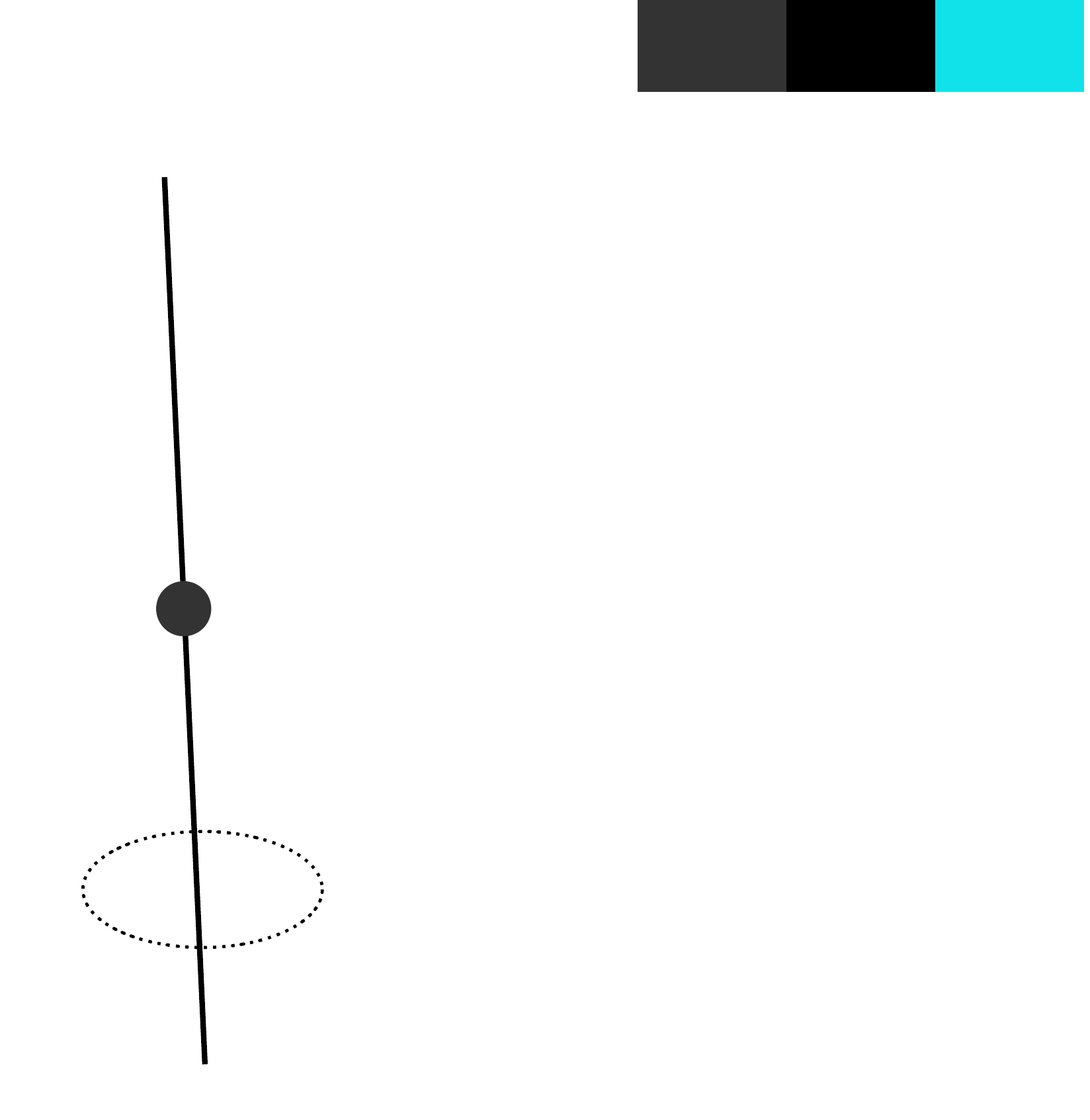
}%
\hspace{3mm}
\subfigure[]{
	\def\svgwidth{0.45\columnwidth}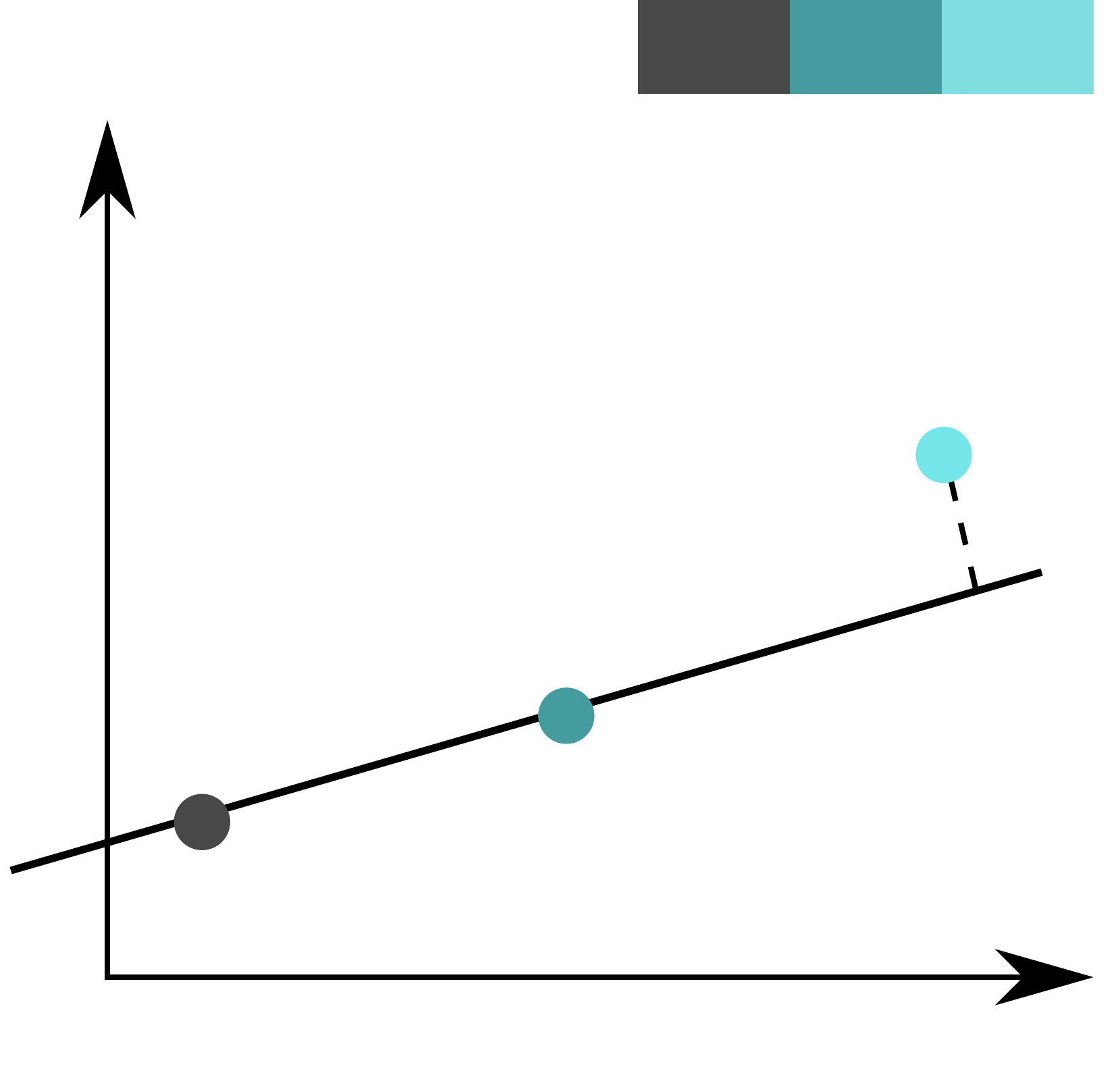
}
\caption{Evaluating Tone Harmony (a) the most vivid color  (turquoise) is not harmonic with the other two  because it does not follow a line (b) Three harmonic points can de described by a line, i.e., the turquoise point is in the vicinity of the previous line.}
\label{fig:threecolors}
\end{figure}

\noindent The contribution of this paper is threefold:%
\begin{description}
\item[Uncertainty of colors.]   
A normal distribution is used to represent uncertainty of colors both on the hue wheel and in the plane of tones. The formulation presented in \cite{lindbloom1994delta} is the basis to obtain an estimation of uncertainty. This representation allows us to compare the hue of colors for evaluating a given pattern\G{adjacency, opposite, or triad}in the color circle, or to compare color  for evaluating  their distance in the tone plane\g{e.g., \mbox{Fig. \ref{fig:twocolors}} illustrates a minimum distance constraint for colors with low chroma and lightness}

\item[Neutral colors.]   Neutral colors have small chroma; i.e., they are located in the vicinity of white-black axis for any cylindrical representation; hence, they usually have inconsistencies in hue. Traditional approaches  tackle this problem by treating near neutral colors as a special case; whereas, the proposed approach associates a large uncertainty in the hue of neutral colors, which allows to use them  without a special treatment. Indeed, this approach is able to combine neutral colors to other colors.
 
\item[Harmonization in the chroma-lightness plane.] The normal distribution associated with each color in the plane of tones also allows to evaluate if the colors approximately follow some geometric pattern. This concept is illustrated in \mbox{Fig. \ref{fig:threecolors}}, first a line joining tones of the first two colors is calculated, if the following point (turquoise) is harmonic to the previous points it must be in the vicinity of the line. 
The line primitive is simple but flexible enough to be adapted to different applications. Uncertainties of points can be propagated to estimate the line uncertainty; hence, the approach can be used incrementally, which is an advantage when selecting one color at a time. The approach also considers that certain line directions are preferred over others.
\end{description}


The rest of this paper is organized as follows: First, Section \ref{sec:related} reviews some of the color theories related to the geometric approach presented in this paper. Then, Section \ref{sec:proposed} explains the uncertainty representation, and its use to discover hue an chroma-lightness patterns. Section \ref{sec:methods} describes the experimental methods used to examine the proposed technique, and Section \ref{sec:related} describes and discusses the results of the study. Finally, Section \ref{sec:results} concludes with a summary of the approach presented and the future areas of interest for this research work on color harmony.

\section{Related Work}\label{sec:related}
Plenty of theories of color harmony have been proposed in the literature, 
as summarized by Westland \etal	\cite{westland2012colour}. The following paragraphs describe some of these theories and their relationship to the geometric approach  presented here.



\begin{figure}[t!]
	\centering
	\def\svgwidth{0.9\columnwidth}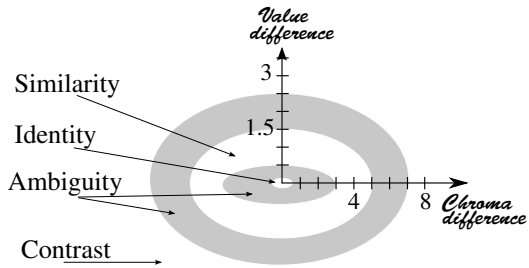
	\vspace{4mm}
	\caption{Difference between two colors in the chroma--ligthness plane (for a given hue). Harmonic regions are interleaved by `ambiguous' regions (gray).}
	\label{fig:moon}
\end{figure}

Moon and Spencer \cite{moon1944geometric} postulate  that  ``harmonious combinations are obtained when:
(i) the interval between any two colors is unambiguous,  and (ii) colors are so chosen that the points representing them in a (metric color) space are related in a simple geometric manner''. The basic aesthetic principle behind the first postulate is that ``the observer should not be confused by the stimuli''; i.e., two harmonic colors not be so close together that there is doubt as to whether they were intended to be identical or only similar. As shown in Fig. \ref{fig:moon}, Moon and Spencer's model considers that harmonic colors give sensations of identity, similarity or contrast. The basic aesthetic principle behind the second postulate is that pleasure is experienced by the recognition of \textit{order}. They suggest several patterns to order colors \g{e.g., points with constant hue must be on a: straight line, circle, triangle, or rectangle}{.}

The relationship between  harmony and color difference is described by Chuang \etal \cite{chuang2001influence} as a cubic function  in which it is possible to identify intervals related to the harmonic regions in the Moon and Spencer's model;  the study also reveals that the first ambiguity color interval is the most critical,  as the perceived color harmony is substantially lower than the seen in the other intervals.   In this paper two close points in the chroma-lightness plane are considered not harmonious \mbox{(Fig. \ref{fig:twocolors}),} as it is difficult to recognize identical colors \g{e.g., color of an object becomes different from one picture to another even with small changes in ambient illumination}{.}

On the contrary, Schloss \etal \cite{schloss2011aesthetic} show that pair preference \G{defined as how much an observer likes a given pair of colors as a Gestalt, or whole} decreases monotonically as a function of the difference in hue. In this paper two hues with similar values are considered harmonious. Uncertainty associated to color values presented here allow explicit representation of the degree of similarity between colors. 

\begin{figure}[t!]
	\centering
	\subfigure[]{
		\centering
		\mbox{
			\def\svgwidth{0.9\columnwidth}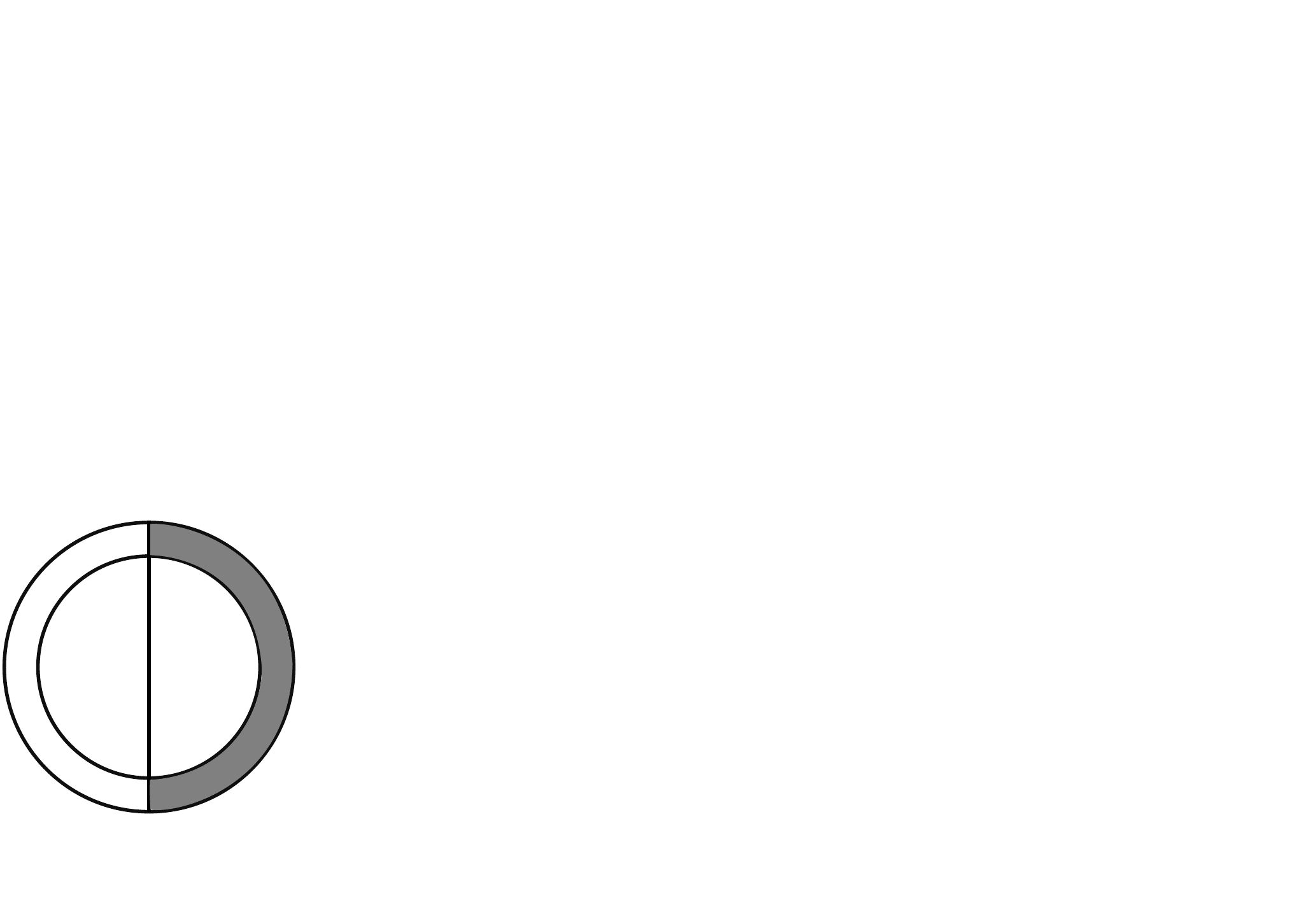
		}
	}\vspace{6 mm}
	\subfigure[]{
		\centering
		\mbox{
			\def\svgwidth{0.9\columnwidth}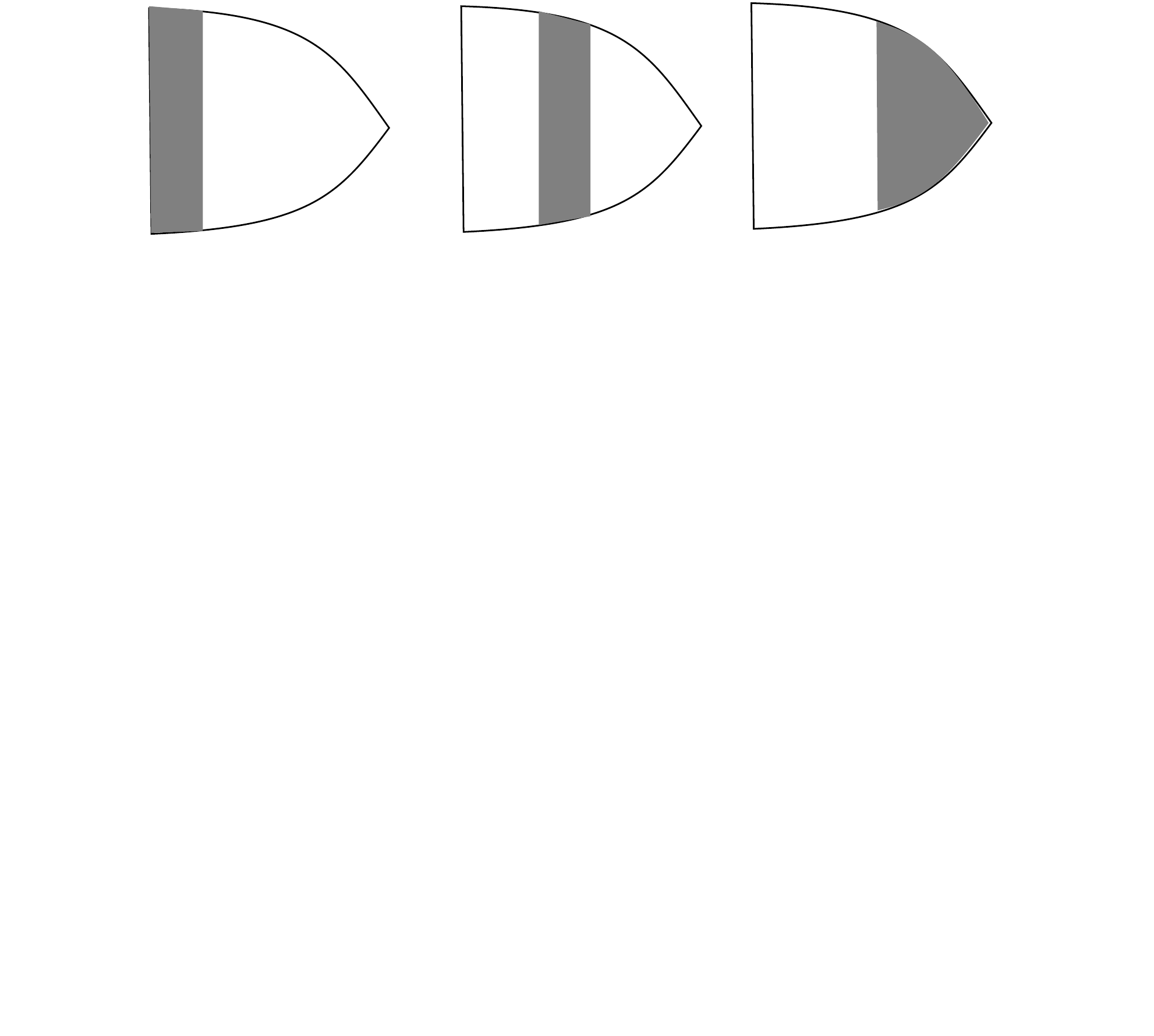
		}
	}
	\vspace{4mm}
	\caption{Matsuda's harmonic templates in Munsell Color Space \cite{matsuda1995color} defined:  (a) on the hue wheel, (b)  in the value and chroma plane.}
	\label{fig:matsuda}	
\end{figure}

Another widely used approach, proposed by Matsuda \cite{matsuda1995color},  suggests that harmonic colors in clothes follow certain patterns both on the hue wheel  (Fig. \ref{fig:matsuda}a) and in the tone plane (Fig. \ref{fig:matsuda}b). We state that the majority of the Matsuda's hue patterns\G{except the Type L and T}can be represented by just three patterns\G{analog, opposite, and triad}with some degree of uncertainty; in addition, a satisfactory description of uncertainty allows to combine `Type N' pattern with other types; e.g., combining neutral colors with lighter or brighter colors, which is a common practice in fashion and design.

Analogously, all the value and chroma patterns proposed by Matsuda\G{except the maximum contrast pattern}follow in some extent a line. This idea also reproduces the second postulate of Moon and Spencer \cite{moon1944geometric} that harmonic colors can be represented in a simple geometric manner. Moreover, we state that some of these linear patters are preferred over others; for instance, Nemcsics \cite{nemcsics2007experimental} examined the variations in the extent of harmony content for compositions that follow lines in the Coloroid saturation--brightness plane. He found that a linear  composition is more harmonic when its angle is  between $30^{\circ }$ and $155^{\circ }$ measured counterclockwise from the positive horizontal line perpendicular to the gray axis. When comparing vertical and horizontal lines for a given hue, he found that vertical lines with saturation between 25 and 45 and horizontal lines with brightness between 75 and 45 are felt more harmonic.

There are several authors that propose to manage color preferences by representing uncertainty; for instance, Hsiao \etal \cite{hsiao2008computer} integrates fuzzy set theory to build an aesthetic measure based color design/selection system.  Lu \etal \cite{lu2015towards}  categorize models to generate color harmony as:  \mbox{(i) \textit{empirical}} (defined by artists and designers), and (ii) \textit{learning--based} methods (those that use machine learning); then they formulate a Bayesian color harmony model where empirical methods are used as the prior,  and the patterns discovered from learning-based methods as the likelihood. One key advantage of the approach proposed here is that it does not requires the outcomes of machine-learning techniques and the uncertainty closely follows the widely accepted color difference formula \cite{sharma2005ciede2000}.

\section{Uncertain Harmonic Patterns}\label{sec:proposed}

This section first describes the uncertainty estimation of each color which is based on the color difference formula \cite{sharma2005ciede2000}, then each color is used to evaluate patterns in the hue circle and geometric pattern in the chroma-lightness plane.

\subsection{Color Uncertainty}
Given a color point $\CMcal{C} = (L,c,h)$ in CIE $L^*c^*h$ coordinates, with lightness $L \in [0,100]$,  chroma $c \in [0,100]$, and hue $h=[0^{\circ },360^{\circ })$, Henceforth, we refer as hue of color by $h$ and the tone of a color  $\CMcal{C}= (L,c,h)$  as%
\begin{equation*}
t(\CMcal{C} ) = t = (L,c).
\end{equation*}

\paragraph{Hue variance.} %
The standard deviation of hue is approximated as: 
\begin{eqnarray}
\nonumber \sigma_{h} &=& f(h,c)\\
 \nonumber &=& k_hS_h + k_N \frac{\gamma^2}{c^2 + \gamma^2} \\
 & =& k_h(1 + 0.015 \; c \; H_T) + k_N \frac{\gamma^2}{c^2 + \gamma^2}
\label{sigmaH}
\end{eqnarray}%
where $k_h, k_N$ and $\gamma$ are parameters, and $H_T$ makes the hue space more uniform, 
\begin{eqnarray*}
	H_T & = & 1-0.17\cos(h-30^{\circ })+0.24\cos(2h)+ \\
	& + & 0.32\cos(3h+6^{\circ })-0.20\cos(4h-65^{\circ }).
	\end{eqnarray*}%
%
The value of $\gamma$ in the second term of \eqref{sigmaH} controls the set of colors perceived as `neutral'. Henceforth, hue of $\CMcal{C}$ is modeled as a normal distribution:%
\begin{equation}
H \sim \mathcal{N}(h, \sigma_h^2) = \mathcal{N}(h, f^2(h,c)).
\label{eq:hueDistr}
\end{equation}%
%


\paragraph{Chroma-lightness Covariance.} %
The tone of a color, $t(\CMcal{C})$, is modeled as a bivariate normal distribution:%
\begin{equation}
T_j \sim \mathcal{N}([c,L]^\intercal,\Sigma_{cL}).
\label{eq:toneDistribution}
\end{equation}%
where
\begin{eqnarray}
	\Sigma_{cL} =& \left[     
	\begin{array}{cc}
		k_c^2S_c^2 & 0   \\
		0 & k_L^2S_L^2 
	\end{array}
	\right],
	\label{covCL}
\end{eqnarray}%
 $k_C$ and $k_L$ are constants and %
\begin{eqnarray*}
	S_{L}&=& 1+{\frac {0.015\left(L-50\right)^{2}}{\sqrt {20+{\left(L-50\right)}^{2}}}}\\
	S_{c}&=& 1+0.045 \; c.
\end{eqnarray*}        

\subsection{Hue Patterns}

\begin{figure}[t!]
	\centering
	\def\svgwidth{0.9\columnwidth}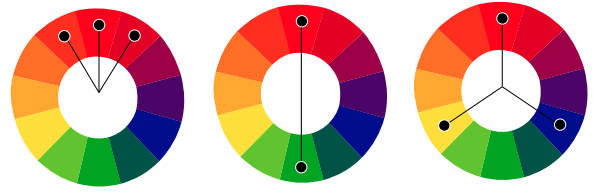

	\caption{Simple patterns in the hue circle (shown from left to right): \textit{Analog}, colors are close to each other; \textit{opposite}, colors are approximately $180^{\circ }$; and \textit{triad}, colors are approximately $120^{\circ }$ each other.}		\label{fig:huePatterns}
\end{figure}

For simplicity, the proposed approach only considers three patterns in the chromatic circle\G{\textit{analog}, \textit{opposite}, and \textit{triad}}as shown in Fig. \ref{fig:huePatterns}. To evaluate them, hue values are standardized and compared.

Let $\theta_m, \theta_n \in [0^{\circ },360^{\circ })$ be two angles, and be $\Delta \theta(\cdot,\cdot)$ the central angle subtended by them. The standardized 
$i$--th difference is defined as %
\begin{equation*}
\Delta^i  \theta (\theta_m,\theta_n) = \frac{\Delta \theta \left(\alpha(\theta_m),\alpha(\theta_n)\right)}{i}
\end{equation*}%
where
$$
\alpha(\theta) = i(\theta \Mod{\tfrac{360}{i})};
$$%
values $i=1,2,3$ are used to discover \textit{analog}, \textit{opposite}, and \textit{triad} patterns, respectively. 

Let $H_j = \mathcal{N}(h_j, \sigma_{h_j}^2)$, $H_k=\mathcal{N}(h_k, \sigma_{h_k}^2)$ be two distributions of the hue values for colors $\CMcal{C}_j$ and $\CMcal{C}_k$, respectively, and $D_B(P,Q, d_{\mu_P\mu_Q})$ the  Bhattacharyya distance of two distributions with $d_{\mu_P\mu_Q}$ a distance of $\mu_P$ and $\mu_Q$.  Colors $\CMcal{C}_j$, $\CMcal{C}_k$ are harmonious on the chromatic circle for the $i$--th pattern if%
\begin{equation}
D_B(H_j, H_k, \Delta ^i \theta(h_j,h_k)) \leq 3.
\label{eq:huecomparison}
\end{equation}%

\begin{algorithm}[t!]
	\DontPrintSemicolon
	\KwData{$\mathcal{L} = [(L_1,c_1,h_1), \ldots, (L_N,c_N,h_N)]$: a list of $N$ colors.}
	\KwResult{Hue harmony label, $l_H \in {}$ \{0 $\mapsto$ \textit{`no harmonic'},  1  $\mapsto$ \textit{`analog'}, 2 $\mapsto$  \textit{`opposite'}, 3 $\mapsto$  \textit{`triad'}\} }
	Calculate hue distribution $H_i$ using \eqref{eq:hueDistr} for each color \;
	\For{$i\in \{1,\ldots, 3\}$\label{op0}}{
		$\hat{H} \gets \mathcal{N} (h_1, \sigma_{h_1}^2)$\;
		\For{$j\in \{2,\ldots, N\}$}{	
			\eIf{$\hat{H}$ and $H_j$ are harmonious (Eq.\ref{eq:huecomparison}) for  pattern $i$  }{
				$\hat{H} \gets $ Fuse $\hat{H}$ and $H_j$ with (Eq.\ref{eq:hueFusion})\;
				\lIf{$j = N$}{return $i$ }
				
			}{
			break\;
		}
	}
}
return 0\;
\caption{Evaluating Hue Harmony}
\label{alg:hueIncremental}
\end{algorithm}

To obtain an incremental algorithm, it is necessary to fuse two hue distributions.
Let $\mathcal{N}(h_j, \sigma^2_{h_j})$ and $\mathcal{N}(h_k, \sigma^2_{h_k})$ be two normal hue distributions which are harmonic, the estimated values for hue and chroma could be obtained as:%
\begin{eqnarray*}
\hat{h} &=& \frac{v_j h_j + v_kh_k}{v_j+v_k},\\
\hat{c} &=& \frac{v_j c_j + v_kc_k}{v_j+v_k},
\end{eqnarray*}%
where $v_j = \nicefrac{1}{\sigma^2_{h_j}}$, and $v_k = \nicefrac{1}{\sigma^2_{h_k}}$. By using \eqref{eq:hueDistr}, the new hue  follows a normal distribution:%
\begin{equation}
\hat{H} = \mathcal{N}{(\hat{h}, f^2(\hat{h}, \hat{c}))}
\label{eq:hueFusion}
\end{equation}

Algorithm \ref{alg:hueIncremental} is used to evaluate a list $\mathcal{L}$ of colors. The input list $\mathcal{L}$ can be sorted according to certain criteria\g{e.g., area of each color}{.} Although any harmony label can be assigned, the algorithm prioritizes `analog' above `opposite'  and  `opposite' above `triads' (line \ref{op0}); as a consequence, the algorithm follows the Occam's razor principle from philosophy: Suppose there exist two explanations for an occurrence, the simpler one is usually better.

\subsection{Chroma-Lightness Patterns}

This section is devoted to describe models for evaluating spatial relations of colors in the chroma--lightness plane.  As stated earlier,  two harmonic colors must not have the same tone (Fig. \ref{fig:twocolors}). Instead of the color difference formula \cite{sharma2005ciede2000}, the  Bhattacharyya distance $D_B$ for multivariate distributions is used here to evaluate proximity of tones. Two tones $t_i$, $t_j$ with distributions $T_i$ and $T_j$ are considered not ambiguous if%
\begin{equation}
D_B(T_i, T_j) \geq 3.
\label{eq:similarTones}
\end{equation} 

This paper explores the hypothesis that harmonic colors follow a line in the chroma--lightness plane; hence, the following paragraphs introduces the estimation of parameters and covariance of a line.

\begin{figure}[!tb]
	\centering
	\def\svgwidth{0.7\columnwidth}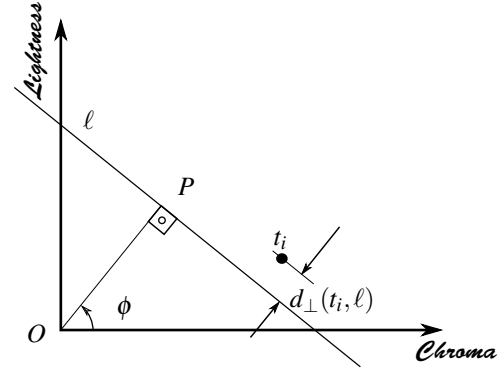
	\caption{Line parameters in the polar form. The shorter distance from the origin to the line $\ell$ is  $r =\left\lvert OP\right\rvert$. }
	\label{fig:lineParams}
\end{figure}

 A line in the plane is represented by its normal form,
$\ell = \params{r, \phi}$; 
where $r$ and $\phi$ are the length and the angle of inclination of the normal, respectively. As shown in Fig. \ref{fig:lineParams}, the normal is the shortest segment between the line and the origin of a given coordinate frame.  %
Points $t = \params{c,L}$ on the line  $\ell = \params{r, \phi}$  satisfy  
$r_j =  c \cos \phi_j  + L \sin \phi_j$.

Given a set of points $D = \set{t_i = \params{c_i, L_i}  | i = 1, \ldots,	 n}$, the maximum likelihood line $\hat{\ell} = \params{\hat{r}, \hat{\phi}}$,
in the weighted least--squares sense \cite{Draper1998}, 
is calculated as
\begin{eqnarray}
\hat{r} & =&  \overline{c}  \cos \hat{ \phi } + \overline{L} \sin \hat{ \phi } \nonumber \\
\hat{\phi} &=& \frac{1}{2} \arctan \frac{-2\sum_{i} w_i  (\overline{L}-L_i)    (\overline{c} -c_i)}%
{\sum_{i} w_i [(\overline{L} -L_i)^2-(\overline{c}-c_i)^2]}, \label{eq:tls}
\end{eqnarray}
where $\overline{c} =  (\sum w_i c_i ) /  ( \sum w_i ) $ and $\overline{L} =  (\sum w_i L_i ) / ( \sum w_i)  $, are the weighted means; as stated by \cite{Siegwart:2011} the individual weight $w_i$ for each measurement can be modeled as $w_i =  (k_cS_ck_L S_L)^{-2}.$

The covariance of the line  is%
\begin{equation}
C_\ell = \sum_{i=1}^n b_iC_ib_i^\intercal
\label{eq:lineCov}
\end{equation}
where $C_i$ is the covariance of the $i$-th point, and $b_i$ is the Jacobian matrix defined as%
\begin{equation*}
b_i = \begin{bmatrix}{\partialfx{\hat{r}}{c_i}}&{\partialfx{\hat{r}}{L_i}}\\
	\partialfx{\hat{\phi}}{c_i}&{\partialfx{\hat{\phi}}{L_i}}\end{bmatrix}.
\end{equation*}

The distance of a tone $t=(c,L)$ to a given line $\ell = \params{\hat{r}, \hat{\phi}}$ is%
\begin{equation*}
d_\perp = \left|  \hat{r} - c \cos \hat{\phi} - L \sin \hat{\phi} \right|;
\end{equation*}%
hence, the variance of the distance is %
\begin{equation*}
\sigma^2_{d_\perp} = J \cdot
\begin{bmatrix}
C_i & 0 \\
0 & C_\ell
\end{bmatrix}
\cdot J^\intercal,
\end{equation*}%
where $ J =
[\partialFx{}{\hat{r}}d_\perp, \partialFx{}{\hat{\phi}} d_\perp,\partialFx{}{x}d_\perp, \partialFx{}{y} d_\perp ]
$. Finally, the point is inlier of $\ell$ if%
\begin{equation}
d_\perp - 2\sqrt{\sigma^2_{d_\perp}} \leq t_\ell.
\label{eq:inlier}
\end{equation}

Algorithm \ref{alg:toneIncremental} discovers tone patterns in an incremental way. First, it calculates the tone distribution of each color in the sorted list $\mathcal{L}$ (line \ref{step:toneDst}). The set $\mathcal{T}$ includes harmonic tones, first it only has the element $T_1$ (line \ref{step:initT}). Then the rest of colors in the list $\mathcal{L}$ are analyzed one-by-one in sequence (line \ref{step:seq}). Two or more harmonic tones are separated
enough from each other (line \ref{step:separated}) and form a geometric line (line \ref{step:line}). Once a tone is accepted as harmonic, it is included in the list (line \ref{step:add}) and the parameters and covariance of the line are updated (lines \ref{step:tls} and \ref{step:lCov}).

\begin{algorithm}[t!]
	\DontPrintSemicolon
	\KwData{$\mathcal{L} = [(L_1,c_1,h_1), \ldots, (L_N,c_N,h_N)]$: a list of $N$ colors.}
	\KwResult{Tone harmony label, $l_t \in {}$ \{0 $\mapsto$ \textit{`no harmonic'},  1  $\mapsto$ \textit{`point'}, 2 $\mapsto$  \textit{`line'}\} }
	Calculate tone distribution $T_j$ using
	\eqref{eq:toneDistribution} for each color \label{step:toneDst}\;
	
	\lIf{$N=1$}{return 1 }
	
	$\mathcal{T} \gets \set{T_1}$ \label{step:initT}\;

		\For{$j\in \{2,\ldots, N\}$ \label{step:seq}}{	
			\If{ $\exists T_i \in \mathcal{T} \; \text{ambiguous with } T_j$ (Eq.\ref{eq:similarTones}) \label{step:separated}}{ return 0}
			\eIf{$(j=2) \vee (T_j$ is inlier of line $\ell$ (Eq.\ref{eq:inlier})$)$ \label{step:line}}
				{$\mathcal{T} \gets \mathcal{T} \cup \set{T_j}$ \label{step:add}}
				{\Return 0}
			$\ell \gets $ Estimate line with the set 	$\mathcal{T}$ (Eq. \ref{eq:tls}) \label{step:tls}\;
			$C_\ell \gets $ Estimate covariance of $\ell$ (Eq.\ref{eq:lineCov})\label{step:lCov};
			
}
return 2\;
\caption{Evaluating tone harmony}
\label{alg:toneIncremental}
\end{algorithm}



\section{Methods}\label{sec:methods}
Given a set of colors in the CIELCh model that follow the proposed uncertain geometric patterns, the purpose of this study is twofold:  %
first, to assess the minimum conditions to obtain color harmony  \g{in terms of hue patterns and the non-ambiguos condition  \eqref{eq:similarTones}}{.} Once these conditions has been established, the second objective is to assess the effect of line's parameters on the color's harmony and to compare colors that do or do not follow the proposed linear patterns.

\begin{figure}[!t]
	\centering
	\def\svgwidth{0.35\columnwidth}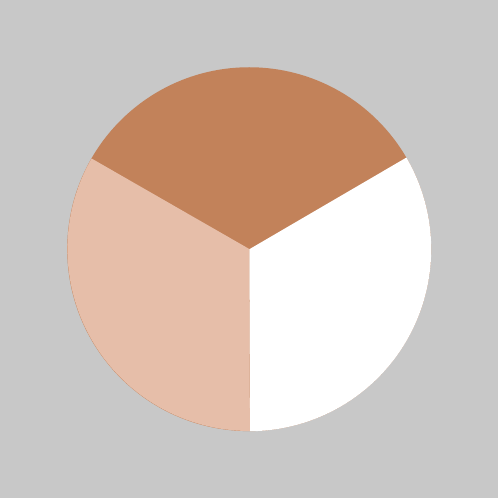\vspace{4mm}
	\caption{Circular pattern of three colors used to show combinations.}
	\label{fig:graficaMetodo}
\end{figure}

\parr{Displaying Color Combinations}
%
The circle shown in Fig. \ref{fig:graficaMetodo}  was used to display combinations of three  colors, that circle is placed on a neutral gray background color. Although the proposed approach can be used for any number of colors, the study  compares combinations of three colors because they are easier to generate and evaluate. Participants were asked to rate each combination on a 10-point Likert scale ranging from `inharmonic' to `harmonic'.

\parr{Subjects}
Thirty  volunteers participated in this study, fifty in each experiment (age range 23-35, Experiment 1; 21-38, Experiment 2).

\parr{Minimum conditions (Experiment 1) }
This study was designed to assess hue patterns and the non-ambiguous condition \eqref{eq:similarTones} of the parameters of the chroma-lightness line on the harmony perception.
Algorithm  \ref{alg:generateLines} was used to generate different color combinations  by varying the parameters
of the line $\params{r, \phi}$ within reasonable bounds.

\parr{Comparison of linear and non-linear patterns (Experiment 2)}
This study was designed to compare the harmony perception of combination of colors that do or do not follow  a linear pattern. Algorithm  \ref{alg:generateLines}  was used to generate different color combinations  by varying the parameters of the line $\params{r, \phi}$ within reasonable bounds. Lines were generated by setting $k_c = k_L = 2$. For each  single linear combination,  we create a corresponding a non-linear combination by simply moving one of the three colors.  Combinations that do not acomplish the minimum conditions established in Experiment 1 were excluded from this experiment. The position of showing linear and non-linear patterns was assigned randomly.

\begin{algorithm}[t!]
	\DontPrintSemicolon
	\KwData{$\params{r, \phi}$: Parameters of the testing line $\ell$,\\ $k$:  Number of colors to be tested.}
	\KwResult{A list of $k$ colors for testing. If the combination of colors was not found, then the testing list is empty.}

	\If{there are $k$ points  $p_i = \params{x_i,y_i}$, on $\ell$  such that  $x_i\in [0,100]$,  $y_i\in [0,100]$, and $\forall i \neq j, d(p_i,p_j)\geq 20$}{
	Draw a random value, $h \in [0, 360)$,   from a uniform distribution.
	Draw a random pattern for hue, $u$,  from the following discrete distribution, \pp{analog}{0.3}, \pp{opposite}{0.3},  \pp{triad}{0.1}, and
	\pp{incomplete triad}{0.3}.  \;
	Calculate  hue values, $h_i, i=1,\dots, k$, by using $h$ and $u$.  These hue values  are  randomly
	disturbed {by a normal error with mean zero and standard deviation  calculated from \eqref{sigmaH} with $c_i=x_i$}.\;
	Determine the  color point  $\CMcal{C}_i$ in CIE $L^*c^*h$ and its covariance $\Sigma_{cL}$. First, $\CMcal{C}_i$ is the closest point to color $\CMcal{C}^*_i = (y_i, x_i, h_i)$, and $\Sigma_{cL}$ is calculated {with \eqref{covCL}} .  If the Mahalanobis distance between $\mathcal{N}(t(\CMcal{C}_i), \Sigma_{cL})$  and $p_i$ is less than a threshold, $\CMcal{C}_i$ is included in the testing list $\mathcal{L}$.\;
	\lIf { $|\mathcal{L}| = k$ and $\forall i \neq j, \CMcal{C}_i, \CMcal{C}_j \in \mathcal{L}$ are unambiguous \eqref{eq:similarTones}}{
		\Return  $\mathcal{L}$
	}
	
}{
\Return $[]$\;

}
	\caption{Generate a test list of $k$ colors that follow a given line in the tone plane}
	\label{alg:generateLines}
\end{algorithm}


\begin{figure}[!t]
	\centering
	\def\svgwidth{0.95\columnwidth}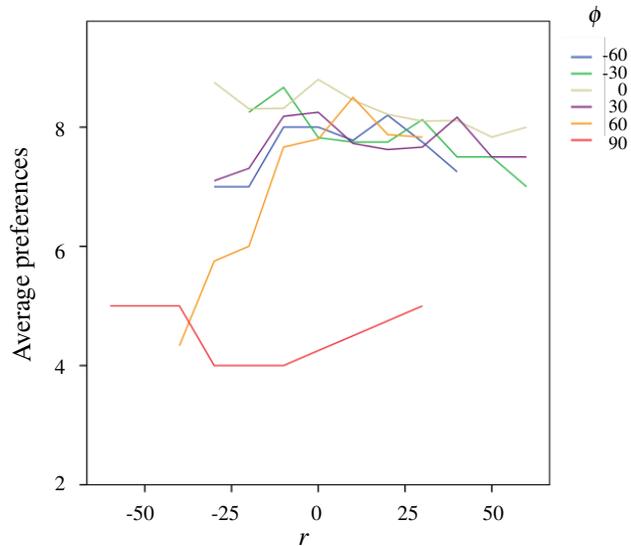\vspace{4mm}
	\caption{Average preferences for different inclinations, $\phi$. Reference point for lines is at $c=20$, $L=60$, angles are in degrees, measured  counterclockwise from the horizontal positive axis. }
	\label{fig:resultLines}
\end{figure}

\section{Results}\label{sec:results}

\parr{Results of Experiment 1}
A one-way between subjects ANOVA was conducted to compare the effect of the hue patterns. There was a significant effect on color preferences at the $p<0.05$ level for the four hue patterns [F(3, 582) = 5.231, $p = 0.001$].  Post hoc comparisons using the Tukey HSD test indicated that the mean score preference for triad (M=6.81, SD=1.14)  was significantly different than the `analog' (M=7.72, SD=1.14) and `incomplete triad' (M=7.61, SD=1.24), However, the `opposite' pattern (M=7.46, SD=1.28)  do not differ from the other patterns.

There was a  extremely significant difference in the preferences for  combinations that  do acomplish the  non-ambiguous condition ($7.59  \pm 1.2$) in comparison to combinations that do not   ($6.07  \pm 1.81$), \mbox{t(1026)= 16.18,} $p < 0.005$.


\parr{Results of Experiment 2}
There was a  extremely significant difference in the preferences for  combinations that  follow a linear pattern  ($7.82  \pm 1.036$) and combinations that do not   ($6.02 \pm 1.36$), \mbox{t(388)= 31.065,} $p < 0.005$.  Preferences according to the angle of inclination of the line are shown in Fig. \ref{fig:resultLines}. Lower preferences for lines at $90^{\circ }$ were observed in comparison to  lines at other inclinations. A similar effect was observed for lines at $\phi = 60^{\circ }$ and $r < -25$.

\section{Conclusion and Future Work}\label{sec:conclusion}
Results show that the proposed uncertain harmonic patterns can generate harmonic combinations.  The non-ambiguous condition \eqref{eq:similarTones} is mandatory to generate harmonic combinations. Results also show that linear patterns \eqref{eq:similarTones} outperform other patterns. 

Color harmony depends on many intrinsic factors \g{e.g., a person may consider certain harmonious colors based on their culture, age, social status, etc}{.} On the other hand, what is considered ugly or nice depends on extrinsic factors like climate, type of emotion that the user wants to transmit, fashion, etc. The difference in preference mean for linear patterns found between experiments 1 and 2, can be explained by these intrinsic  (unhomogeneous groups) and extrinsic (a contrast effect caused by showing two similar patterns one after  other.

We are developing an App that helps people to generate harmonic combinations for clothing. Hence,  more accurate information can be obtained from the users' preferences.  Besides completing our theory we are adapting a clothing search system to user's subjectivity.



\phantomsection



\end{document}